\documentclass{article}

\PassOptionsToPackage{numbers, compress}{natbib}

\usepackage[final]{nips_2018}

\usepackage[utf8]{inputenc} 
\usepackage[T1]{fontenc}    
\usepackage{hyperref}       
\usepackage{url}            
\usepackage{booktabs}       
\usepackage{amsfonts}       
\usepackage{nicefrac}       
\usepackage{microtype}      
\usepackage{mwe}
\usepackage[export]{adjustbox}

\title{Image Classification at Supercomputer Scale}

\author{
    Chris Ying, Sameer Kumar, Dehao Chen, Tao Wang, Youlong Cheng \\
    Google, Inc. \\
    \texttt{\{chrisying, sameerkm\}@google.com}
}

\begin{document}

\maketitle

\begin{abstract}
Deep learning is extremely computationally intensive, and hardware vendors have responded by building faster accelerators in large clusters.
Training deep learning models at petaFLOPS scale requires overcoming both algorithmic and systems software challenges.
In this paper, we discuss three systems-related optimizations:
(1) distributed batch normalization to control per-replica batch sizes,
(2) input pipeline optimizations to sustain model throughput,
and (3) 2-D torus all-reduce to speed up gradient summation.
We combine these optimizations to train ResNet-50 on ImageNet to 76.3\% accuracy in \textbf{2.2 minutes} on a 1024-chip TPU v3 Pod with a training throughput of over \textbf{1.05 million images/second} and \textbf{no accuracy drop}.
\end{abstract}

\section{Introduction}

The success of deep neural networks in a wide range of applications has been driven by dramatic increases in computational power.
At the core, the vast majority of neural networks use some variant of stochastic gradient descent (SGD) to optimize model weights.
Models typically require multiple SGD passes through a dataset in order to converge.
This results in a substantial number of floating point operations to be executed for the model to converge.
For example, on ImageNet [\citenum{ILSVRC15}], the commonly benchmarked ResNet-50 [\citenum{DBLP:journals/corr/HeZRS15}] model executes on the order of $3.2 \times 10^{16}$ floating point operations for a single epoch.
Typical training regimes require up to 90 epochs to converge.

While hardware accelerators such as GPUs and TPUs have sped up the iteration times, large neural networks can still take hours or days to train large datasets on a single accelerator.
One commonly used technique to speed up training is parallelization across multiple devices via distributed SGD where the mini-batch is distributed among many replicas.
Historically, asynchronous distributed SGD was used to train across a large number of workers [\citenum{dean2012large}], but recent work [\citenum{DBLP:journals/corr/ChenMBJ16}] has shown that asynchronous SGD does not perform as well as synchronous SGD in terms of time to convergence and final achieved validation accuracy.
Even when using synchronous distributed SGD, large-scale training requires standard machine learning techniques to be adapted to maintain model quality.
Furthermore, specialized systems optimizations are necessary to achieve real-world end-to-end wall clock speed-up.

Some of the challenges which we observe are:
\begin{enumerate}
\item Model accuracy depends on the global batch size and the per-replica batch size.
\item Training becomes bottlenecked by CPU input processing when the accelerator computation becomes sufficiently fast. 
\item Performing synchronized distributed gradient descent requires a highly scalable all-reduce implementation.
\end{enumerate}

This paper presents the systems-related optimizations we utilized to address the above challenges. In section \ref{sec:methods}, we introduce each system optimization we used. In section \ref{sec:analysis}, we use controlled experiments to show the impact of each individual optimization. In section \ref{sec:results}, we put together all the optimizations as well as machine learning techniques to train ResNet-50 in state-of-the-art wall clock time.

\subsection{Tensor Processing Units}

\begin{figure}
    \centering
    \begin{minipage}{0.45\textwidth}
        \centering
        \includegraphics[width=0.9\textwidth]{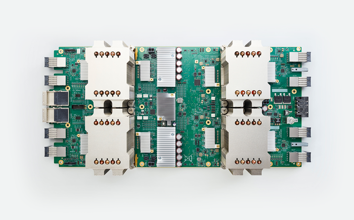}
        \caption{Cloud TPU v2 device with four chips, \textbf{180 teraFLOPS} of peak floating point throughput and \textbf{64 GB} of High Bandwidth Memory (HBM).}
        \label{fig:tpuv2}
    \end{minipage}\hfill
    \begin{minipage}{0.45\textwidth}
        \centering
        \includegraphics[width=0.9\textwidth]{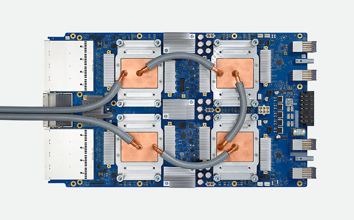}
        \caption{Liquid-cooled Cloud TPU v3 device with four chips, \textbf{420 teraFLOPS} of peak floating point throughput and \textbf{128 GB} of HBM.}
        \label{fig:tpuv3}
    \end{minipage}
\end{figure}

In this paper, we ran our experiments on Google Tensor Processing Unit (TPU) accelerators [\citenum{Jouppi:2017:IPA:3079856.3080246}].
However, we emphasize that the techniques presented here are not limited to TPUs and can be applied to other architectures as well.
The TPU chips have matrix processing units to accelerate dense matrix multiplication operations in convolutional neural networks.
Cloud TPU v2 accelerator specifications can be found in Figure \ref{fig:tpuv2}, and Cloud TPU v3 accelerator specifications can be found in Figure \ref{fig:tpuv3}.
Each TPU device has four chips and each chip has two separate cores.

TPU v2 devices can be connected together in a 256-chip configuration, called a TPU Pod, with a total of 11.5 petaFLOPS of mixed-precision throughput. The TPU v3 Pod is four times larger than the TPU v2 Pod, containing 1024 chips to deliver a theoretical peak of 107.5 petaFLOPS of mixed-precision throughput.

\section{Related Work}

Large-scale deep learning, particularly in the context of large-batch training, has been explored before. \citet{DBLP:journals/corr/KeskarMNST16} observed that there is a "generalization gap" between models trained at small and large batches and explain that the gap may be due to large batch training converging to a sharp minima that does not result in good validation accuracy.
\citet{DBLP:journals/corr/GoyalDGNWKTJH17} trained ResNet-50 on ImageNet with 8192 batch size in one hour via a combination of learning rate scaling and learning rate warmup techniques.
\citet{DBLP:journals/corr/abs-1708-03888} tuned the per-layer learning rate, based on the ratio of the norm of the weights and the gradients, to scale ResNet-50 to up to 32768 batch size. 
\citet{DBLP:journals/corr/abs-1711-00489} demonstrated an alternative way to deal with the generalization gap by increasing the batch size during the training schedule in addition to the learning rate.
\citet{DBLP:journals/corr/abs-1711-04325} used a combination of batch normalization changes and optimizer/learning rate schedule changes to train ResNet-50 in 15 minutes.
\citet{jia2018highly} further reduced ResNet-50 train time to 8.7 minutes with peak validation accuracy via mixed-precision training and optimized all-reduce algorithms.
\citet{DBLP:journals/corr/abs-1708-07120} approached the problem from a different angle by reducing the number of iterations rather than increasing the throughput.

\section{Methods} \label{sec:methods}

Based on prior related work in large-batch training, we utilize the following techniques in our experiments:

\paragraph{Mixed precision} In our experiments, convolutions are executed using bfloat16, a 16-bit half precision floating point format on TPUs.
Furthermore, input images and intermediate activations are also stored in bfloat16.
To maintain comparable accuracy with 32-bit floating point networks, all non-convolutional operations (e.g. batch normalization, loss computation, gradient summation) use 32-bit floating point numbers.
Since the majority of the computational and memory access overheads are in the convolutional operations, using bfloat16 enables higher training throughput with minimal or no loss in model accuracy [\citenum{jia2018highly}].

\paragraph{Learning rate schedules} Past studies [\citenum{DBLP:journals/corr/Krizhevsky14}] [\citenum{hoffer2017train}] have shown that learning rates should be proportional to the batch size.
In our experiments, we employ linear learning rate scaling (i.e. double the batch size, double the learning rate).
We also utilize gradual learning rate warm-up and learning rate decay.

\paragraph{Layer-wise Adaptive Rate Scaling (LARS)} While simple stochastic gradient descent with momentum allowed us to scale to batch 8192. The LARS optimizer [\citenum{DBLP:journals/corr/abs-1708-03888}], however, enabled scaling further to up-to 32768 batch size with no loss in model quality. The larger batch size of 32768 also improved the model execution throughput on TPUs.

In addition to these prior methods, we discuss three more optimization techniques to enable efficient scaling to hundreds of accelerators.

\subsection{Distributed Batch Normalization}

Batch normalization (BN) [\citenum{DBLP:journals/corr/IoffeS15}] is a critical part training image classification models. In a distributed training setting, a common practice is to perform BN per replica, which reduces the cross-device communication cost. In practice, we've found that the BN batch size (i.e. the per-replica batch size) has a critical effect on the final validation accuracy achieved in the model. It is observed in [\citenum{DBLP:journals/corr/abs-1803-08494}] that when the batch size per replica is below 32, the ResNet-50 model does not converge to peak validation accuracy. When scaling up to a very large number of workers via data parallelism, either the global batch size must be scaled up linearly or the per-replica batch size must be scaled down. Prior work [\citenum{DBLP:journals/corr/GoyalDGNWKTJH17}] has shown that validation accuracy suffers at larger batch sizes. In the context of BN, we focus on the case where the per-replica batch size is small.

\begin{figure}
  \centering
  \includegraphics[width=0.9\textwidth]{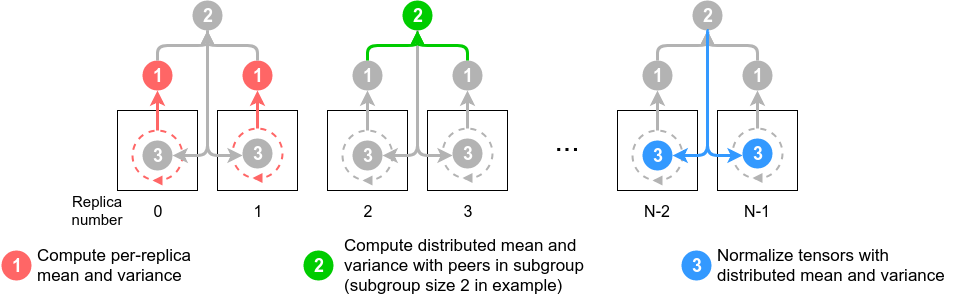}
  \caption{Distributed batch normalization algorithm. In this example, the subgroup size is 2.}
  \label{fig:dbn}
\end{figure}

We enhance BN with a distributed reduction across a small number of peers to compute the mean and variance on a subset of replicas. This gives us control over the BN effective batch size and decouples it from the global batch size and number of replicas. The distributed BN procedure is shown in Figure \ref{fig:dbn}.

\subsection{Input Pipeline Optimization}

The input pipeline to a model includes the filesystem reads, data parsing, preprocessing, shuffling, and batching.
If the input pipeline does not sustain the same throughput as the model pipeline (forward and backwards passes), then the overall training throughput will be bottlenecked by the input pipeline.
The distinction between the input pipeline and model pipeline exists because many parts of the input are inefficient or incompatible with the hardware accelerator and must be performed on CPU, while the model pipeline is completely executed on the device. 

After experimenting with a wide range of techniques, we identified four key optimizations in the input pipeline that stood out for having a significant impact on performance.
Our fastest results would be input-bottlenecked without these following optimizations.
In the context of large scale image classification, we are not aware of prior work combining these techniques.

\paragraph{Dataset sharding and caching} Ideally, a dataset should only be read once and then stored in memory for all further usage during the training procedure.
Practically, this is not very feasible because real-world datasets are often far too large to store in entirely in host memory.
With a large number of workers, it becomes possible to shard the dataset between workers and enable more efficient access patterns to the data.

\paragraph{Prefetch to pipeline input and compute} While the accelerator is training on the current batch, the next batch can be processed on the host input pipeline concurrently, reducing the wait time between the accelerator and data processing on the CPU.
Furthermore, because the ImageNet dataset has images of different sizes, our prefetching creates headroom when processing small images so that it can stay ahead of the device even when it must process larger images.

\paragraph{Fused JPEG decode and cropping} Raw image datasets often come encoded in the JPEG format (for example, ImageNet).
It is common practice to randomly crop [\citenum{DBLP:journals/corr/SzegedyLJSRAEVR14}] and resize the image as part of the preprocessing and data augmentation.
Rather than parsing the original image size, it is more efficient to only decode the relevant part of the encoded image after computing the crop window and that results in significantly lower overheads.

\paragraph{Parallel data parsing} Parsing data and performing the preprocessing is often the most expensive part of the input pipeline.
A multi-core CPU can increase throughput by parallelizing the these operations among several worker CPU threads.

\subsection{2-D Gradient Summation}

\begin{figure}
  \centering
  \includegraphics[width=0.6\textwidth]{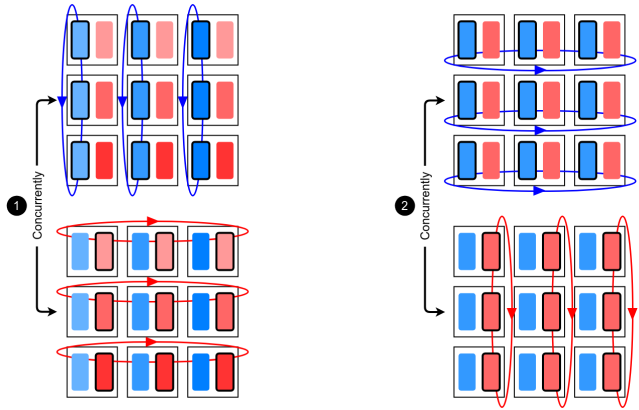}
  \caption{[best viewed in color] 2-D all-reduce across a hypothetical $3 \times 3$ torus. In the first phase (left), the first half of the tensors (blue) are summed along the vertical dimension while the second half of the tensors (red) are summed concurrently along the horizontal dimension. In the second phase (right), the dimensions are flipped, which completes the all-reduce for both halves.}
  \label{fig:2dtorus}
\end{figure}

Prior all-reduce implementations for deep learning [\citenum{baidu-all-reduce}] use a ring-based all-reduce for global gradient summations.
We found that, on a TPU Pod, the 1-D ring all-reduce algorithm was limited by the latency of pushing packets in a Hamiltonian circuit across all the nodes in the pod.
We explored a 2-D mesh algorithm that computes the reduction in two phases, one per dimension.
We utilize two \textit{concurrent} ring reductions each summing \textit{different halves} of the payload along X and Y dimensions (Figure \ref{fig:2dtorus}).
When the TPU network has torus wrap-around links, we also enable bi-directional ring exchange to further speed up the gradient summation operation.
Note, the 2-D algorithm reduces the span of gradient summation to $O(N)$ versus $O(N^2)$ network hops in the 1-D algorithm.
In addition, the 2-D algorithm increases the throughput of gradient summation by a factor of two over the 1-D algorithm. 

\section{Analysis} \label{sec:analysis}

In this section, we perform controlled experiments to demonstrate the individual importance of each of the discussed optimization methods above.

\begin{figure}
    \centering
    \begin{minipage}{0.48\textwidth}
        \centering
        \includegraphics[width=\textwidth]{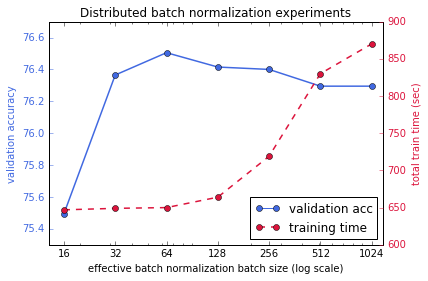}
        \caption{ImageNet validation accuracy and total training time for ResNet-50 for various effective batch normalization sizes. All experiments use 16 per-core batch size, only the distributed group size changes.}
        \label{fig:dbn_exps}
    \end{minipage}\hfill
    \begin{minipage}{0.48\textwidth}
        \centering
        \includegraphics[width=\textwidth]{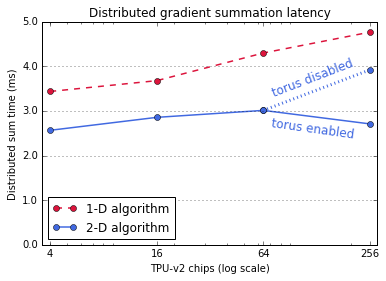}
        \caption{Gradient summation time on ResNet-50 for various chip counts with 1-D vs. 2-D reduction algorithms. Torus links are only available on the full TPU v2 Pod (256 chips).}
        \label{fig:1d2d}
    \end{minipage}
\end{figure}

\paragraph{Distributed Batch Normalization} Figure \ref{fig:dbn_exps} shows a comparison of accuracy and training time for a range of group sizes.
The experiments were conducted on a TPU v2 Pod without using LARS optimizer.
All variables aside from group size/effective BN batch size were controlled.
The baseline of 16 effective BN batch size (group size 1) reaffirms that a per-core batch size below 32 causes a big drop in model performance.
As the group size increases, the cost of computing the group mean and variance increases, which is observed in the increased total training time.
Based on these experiments, we use an effective BN batch size of 64 (group size 4).

\begin{figure}
  \centering
  \includegraphics[width=0.95\textwidth]{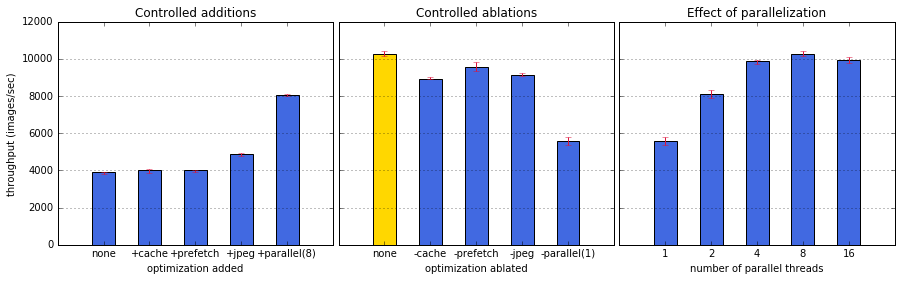}
  \caption{Error bars show 25\% and 75\% quartiles. (left) Controlled additions of each of the 4 optimizations. (center) Controlled ablations of each of the 4 optimizations, the yellow bar indicates the final pipeline that we used which includes all optimizations. (right) Varying the number of parallel threads with all other optimizations enabled.}
  \label{fig:inputs}
\end{figure}

\paragraph{Input Pipeline Optimizations} Based on the four optimizations discussed above (abbreviated: cache, prefetch, jpeg, parallel), we perform controlled additions and ablations (Figure \ref{fig:inputs}).
All experiments are conducted on a single Intel Skylake processor using TensorFlow's \texttt{tf.data.Dataset} API [\citenum{tensorflow2015-whitepaper}].
All experiments are executed over 20 million total images averaged every 1000 images.
These values represent the mean observed throughput per host CPU (in a large-scale training job, there are multiple host CPUs).
From our experiments, parallelization is the most important optimization by itself, more than doubling the base throughput.
The fused JPEG decode and crop also yielded around 25\% improvement when added and around 10\% drop when ablated.
Caching and prefetching show minor improvements when added but cause 6-13\% drop in performance when ablated, implying that the optimizations are crucial at high throughputs.
Note that sharding the ImageNet dataset across a TPU Pod allows us to fully cache the dataset into memory.

\paragraph{2-D Gradient Summation} Figure \ref{fig:1d2d} shows the time to compute gradient reductions in ResNet-50 at various TPU v2 chip counts.
The 2-D algorithm results in more scalable performance as it has lower latency and higher throughput than the 1-D algorithm.
On the 256-chip pod configuration, torus links enable more efficient gradient summation over the baseline 1-D ring and 2-D mesh algorithms.

\section{Results} \label{sec:results}

Having analyzed each optimization in isolation, we now combine the techniques to achieve unprecedented results on real machine learning workloads.
We use the commonly benchmarked ResNet-50\footnote{The ResNet-50 model used here refers to the variant popularized by \citet{DBLP:journals/corr/GoyalDGNWKTJH17}. This version is slightly different than the canonical ResNet-50 v1 [\citenum{DBLP:journals/corr/HeZRS15}] and requires approximately 12\% more compute to train but achieves higher accuracy. We also use label smoothing to improve generalization.} trained on ImageNet.
We intentionally train for 90 epochs to be comparable with past results, though we expect that methods to reduce the epochs necessary will further improve results.
Our top results using TPU Pods are summarized in Table \ref{tab:results}, compared to other state-of-the-art results at the time of publication.
The training time is measured by the time from end of the just-in-time compilation of the TPU binary to the write time of the final checkpoint file. The accuracy is measured on the validation set without removing blacklisted examples.

\pagebreak  

In summary, the techniques we utilize are:
\begin{itemize}
    \item Mixed-precision training using bfloat 16.
    \item Learning rate scaling, warmup, and decay schedule.
    \item LARS optimizer to scale to 32768 batch size.
    \item Distributed batch normalization to control batch normalization batch sizes.
    \item Input pipeline optimizations to sustain the model throughput.
    \item Gradient summation with 2-D algorithm using torus links.
    \item Trained on a 1024 chip TPU v3 Pod using TensorFlow.
\end{itemize}

The systems techniques, including scalable 2-D gradient summation and input pipeline optimization, allowed us to train on 1024 TPU v3 chips in \textbf{2.2 minutes} (including some constant costs at the start of training, e.g. initializing the weights) with over \textbf{1.05 million images/second} training throughput (measured via profiling during training).
Distributed batch normalization in conjunction with LARS enables us to achieve this speed while maintaining 76.3\% accuracy i.e. \textbf{no accuracy drop}.
While our techniques are evaluated on TPUs, they are general and apply to other accelerators as well.

\begin{table}

  \caption{Large-scale ResNet-50 training results.}
  \label{tab:results}
  \centering
  \begin{adjustbox}{center,max width=1.1\textwidth}
  \begin{tabular}{llllllll}
    \toprule                \\
     & Hardware & Chips & Batch & Optimizer & BN & Accuracy & Time \\
    \midrule
    \citet{DBLP:journals/corr/GoyalDGNWKTJH17} & P100 & 256 & 8192 & Momentum & Local & 76.3\% & 1 hour \\
    \citet{DBLP:journals/corr/abs-1711-00489} & TPU v2 & 128 & 8192 $\rightarrow$ 16384 & Momentum & Local & 76.1\% & 30 mins. \\
    \citet{DBLP:journals/corr/abs-1711-04325} & P100 & 1024 & 32768 & RMS + Mom. & Local & 74.9\% & 15 mins. \\
    \citet{jia2018highly} & P40 & 1024 &  65536 & LARS & Local & 76.2\% & 8.7 mins. \\
    \midrule
    Baseline & TPU v2 & 4 & 1024 & Momentum & Local & 76.3\% & 7.2 hours \\
    Ours & TPU v2 & 256 & 16384 & Momentum & Local & 75.3\% & 9.7 mins. \\
    Ours & TPU v2 & 256 & 32768 & LARS & Local & 76.3\% & 8.0 mins. \\
    Ours & TPU v3 & 512 & 32768 & LARS & Local & 76.4\% & 3.3 mins. \\
    Ours & TPU v3 & 1024 & 65536 & LARS & Local & 75.2\% & 1.8 mins. \\
    Ours & TPU v3 & 1024 & 32768 & LARS & Distributed & \textbf{76.3\%} & \textbf{2.2 mins.} \\
    \bottomrule
  \end{tabular}
    \end{adjustbox}
\end{table}

\section{Future Work}

This paper explores scaling image classification using purely data parallelism.
An alternative way to scale up model training is to use a mix of data and model parallelism to shard the computation so that we can utilize large clusters without necessarily using large global batch sizes.


\pagebreak

\subsubsection*{Acknowledgements}

We would like to acknowledge Bjarke Roune and Hyoukjoong Lee for their contributions to distributed gradient summation. We would like to thank Blake Hechtman, David Majnemer, Brennan Saeta, Sourabh Bajaj, Naveen Kumar, Jonathan Hseu and Frank Chen for technical support with compilers and systems libraries, Geroge Dahl for his expertise on large batch training, Yang You for support with the LARS optimizer, and Pieter-Jan Kindermans for help with the ResNet-50 implementation.

\bibliography{bibliography}{}
\bibliographystyle{plainnat}

\end{document}